\documentclass[11pt]{article}

\usepackage{amsfonts,amstext,amsmath,amssymb}
\usepackage{color}
\usepackage{graphicx}
\usepackage{multirow}
\usepackage{rotating}
\usepackage[margin=1in]{geometry}
\usepackage[authoryear,comma,sectionbib]{natbib} 

\newcommand{\E}{\mathbb{E}}
\def\argmin{\mathop{\mathrm{argmin}}}
\def\argmax{\mathop{\mathrm{argmax}}}

\begin{document}

\title{Hidden Markov Models with mixtures as emission distributions}

\author{Stevenn Volant$^{1,2}$, Caroline B\'erard $^{1,2}$, \\
Marie-Laure Martin Magniette$^{1,2,3,4,5}$ and St\'ephane Robin$^{1,2}$}

\date{}

\maketitle
\begin{center}
$^{1}$AgroParisTech, 16 rue Claude Bernard, 75231 Paris Cedex 05, France. \\ 
$^{2}$INRA UMR MIA 518, 16 rue Claude Bernard, 75231 Paris Cedex 05, 
France. \\ 
$^{3}$INRA UMR 1165, URGV, 2 rue Gaston Cr\'emieux, CP5708, 91057, Evry 
Cedex, France. \\ 
$^{4}$UEVE, URGV, 2 rue Gaston Cr\'emieux, CP5708, 91057, Evry Cedex, 
France. \\ 
$^{5}$CNRS ERL 8196, URGV, 2 rue Gaston Cr\'emieux, CP5708, 91057, Evry 
Cedex, France. 
\end{center}

\maketitle
\begin{abstract}

In unsupervised classification, Hidden Markov Models (HMM) are used to account for a neighborhood structure between observations. The emission distributions are often supposed to belong to some parametric family. In this paper, a semiparametric modeling where the emission distributions are a mixture of parametric distributions is proposed to get a higher flexibility. We show that the classical EM algorithm can be adapted to infer the model parameters. For the initialisation step, starting from a large number of components, a hierarchical method to combine them into the hidden states is proposed. Three likelihood-based criteria to select the components to be combined are discussed. To estimate the number of hidden states, BIC-like criteria are derived. A simulation study is carried out both to determine the best combination between the merging criteria and the model selection criteria and to evaluate the accuracy of classification. The proposed method is also illustrated using a biological dataset from the model plant \textit{Arabidopsis thaliana}. A R package \textsc{HMMmix} is freely available on the CRAN.


\end{abstract}

\sloppy



\section{Introduction}

Hidden Markov models (HMM) constitute an efficient technique
of unsupervised classification for longitudinal data. HMM have been
applied in many fields including signal processing \citep{Rabiner89},
epidemiology \citep{Sun2009} or genomics \citep{Li2005bio,Durbin1999}.
In such models, the neighbourhood structure is accounted for via a
Markov dependency between the unobserved labels, whereas the
distribution of the observation is ruled by the so-called 'emission'
distribution. 
In most cases, the emission distributions associated with the hidden states
are given a specific form from a parametric class such as Gaussian, Gamma or Poisson.
This may lead to a poor fit, when the distribution of the data is far from the
chosen class. Efforts have been made to propose more complex
models capable of fitting skewed or heavy-tailed distribution, such as
the multivariate normal inverse Gaussian distribution proposed in
\citep{Chatzis2010}. However the case of multimodal emission distributions has been  little studied.

In the framework of the model-based clustering, where no spatial dependence of the latent variable is taken into account,
a great interest has been recently paid to the definition of more
flexible models using mixture as emission distributions
\citep{Li2005,Baudry2008}. 
This approach can be viewed as semi-parametric as the shape of 
the distribution of each component of these mixtures 
is hoped to have a weak influence on the estimation of
the emission distributions.
The main difficulty raised by this approach is to combine the
components. To achieve this step, classical clustering
algorithms are generally used: $k$-means approach \citep{Li2005} or hierarchical clustering \citep{Baudry2008}. For a general
review on the merging problem of Gaussian components in the
independent case, see \citep{Hennig2010}. To the best of our knowledge, these
approaches have been limited to independent mixture models until now.

In this paper, we propose to extend this semi-parametric modeling to
the HMM context. We first show that the inference can be fully achieved using 
the EM algorithm (Section \ref{Model}).  
Section \ref{Methodology} is dedicated
to the intialization of the  EM algorithm that aims at merging 
components into the hidden states by considering 
 an HMM version of the hierarchical algorithm of
\citep{Baudry2008}. We then consider the choice of the number of hidden states
and proposed three BIC-like criteria (Section \ref{SelectionCrit}). Based on a simulation study presented in Section
\ref{Simul}, the best merging criterion is chosen. Eventually, the proposed 
method is applied to probe classification in ChIP-chip experiment 
for the plant \textit{Arabidopsis thaliana}
(Section \ref{realdata}). An R package \textsc{HMMmix} is freely
available on the CRAN.

\section{Model and inference} \label{Model}
This section describes the collection of models considered to fit 
the data distribution and the E-M algorithm used to estimate the parameter vector of each model.

\subsection{Model}

We assume that the observed data $X = \{X_1,...,X_n\}$, where $X_t \in \mathbb{R}^{Q}$, are modeled with  an HMM. The latent variable $\{S_t\}$ is a $D$-state homogeneous Markov chain with transition matrix $\Pi = [\pi_{dd'}]$ and stationary distribution $q$. The observations $\{X_t\}$ are independent conditionally to the hidden state with emission distribution $\psi_d$ ($d = 1,...,D$):
\begin{eqnarray}
\label{Psi}
 (X_t|S_t=d) \sim \psi_d.
\end{eqnarray}
We further assume that each of the emission distributions $\psi_d$ is itself a mixture of $K_d$ parametric distributions:
\begin{eqnarray}
\label{equpsi}
\psi_d =  
\sum_{k=1}^{K_d} \lambda_{dk} \phi(.;\gamma_{dk}) ,
\end{eqnarray}
where $\lambda_{dk}$ is the mixing proportion of the $k$-th component from cluster $d$ ($\forall k \in \{1,...,K_d\}$, $0 < \lambda_{dk} < 1$ and $\sum_k \lambda_{dk} = 1$) and $\phi(.;\gamma)$ denotes a parametric distribution known up to the parameter vector $\gamma$. We denote 
by $\theta$ the vector of free parameters of the model.

For the purpose of inference, we introduce a second hidden variable $\{Z_t\}_t$ which refers to the component $k$ within state $d$, denoted $(dk)$. According to (\ref{Psi}) and (\ref{equpsi}), the latent variable $\{Z_t\}$ is itself a Markov chain
with the transition matrix $\Omega=\left[\omega_{(dk), (d'k')}\right]$, where
\begin{equation}\label{PIZS}
  \omega_{(dk),(d'k')}= \pi_{dd'} \lambda_{d'k'},
\end{equation}
so the transition between $(dk)$ and $(d'k')$ only depends on the hidden state $S$ at the previous time.

According to these notations, a model $m$ is defined by $D$ and the 
$D$-uplet $(K_1, \ldots, K_D)$ specifying the number of hidden states and the number of components within each hidden state. 
Finally in the context of HMM with mixture emission distributions, 
we consider a collection of model defined by 
\begin{eqnarray*}
\mathcal{M}=\{ m&:=&(D, K_1, \ldots, K_D);\\
 &&1 \leq D ; \forall d K_d \geq 1 \mbox{ with } \sum_{d=1}^D K_d := K \}.
\end{eqnarray*}
In this paper, we leave the choice of $K$ to the user,
provided that it is large enough to provide a good fit.
\subsection{Inference  for a given model} \label{Inference}
This section is devoted to the  parameter vector estimation of a given 
model $m$ of the collection $\mathcal{M}$. The most common strategy for maximum likelihood inference of HMM relies on the Expectation-Maximisation (EM) algorithm \citep{Dempster77, Capp2010}. Despite the existence of two latent variables
$S$ and $Z$, this algorithm can be applied by using 
the decomposition of the log-likelihood
\begin{eqnarray*}
	\log P(X) & = &  \mathbb{E}_X[\log P(X, S, Z)] - \mathbb{E}_X[\log P(S, Z|X)]
\end{eqnarray*}
where $\mathbb{E}_X$ stands for the conditional expectation, given the observed data $X$. \\
The E-step consists in the calculation of the conditional distribution $P(S, Z|X)$ using the current value of the parameter $\theta^h$. The M-step aims at maximizing the completed log-likelihood $\mathbb{E}_X[\log P(X, S, Z)]$, which can be developed as
\begin{eqnarray*}
	\mathbb{E}_X[\log P(X, S, Z)] & = & \mathbb{E}_X[\log P(S)] + \mathbb{E}_X[\log P(Z|S)] \\
	& & + \mathbb{E}_X[\log P(X| S, Z)]
\end{eqnarray*}
where
$$
 \begin{array}{l}
	\displaystyle{\mathbb{E}_X[\log P(S)] = \sum_{d=1}^D \tau_{1d} \log q(S_1=d) + \sum_{t=1}^n \sum_{d, d'=1}^D \eta_{tdd'} \log \pi_{dd'}} \\
	\displaystyle{\mathbb{E}_X[\log P(Z|S)]  =  \sum_{d=1}^D \sum_{k=1}^{K_d} \sum_{t=1}^n \tau_{td} \delta_{tdk} \log \lambda_{dk}} \\
	\displaystyle{\mathbb{E}_X[\log P(X| S, Z)]  =  \sum_{d=1}^D \sum_{k=1}^{K_d} \sum_{t=1}^n \tau_{td} \delta_{tdk} \log \phi_d(X_t, \gamma_{dk})}
\end{array}
$$
denoting
\begin{eqnarray*}
	\tau_{td} & = & P(S_t = d|X) \\
	\eta_{tdd'} & = & P(S_t=d, S_{t+1}=d' |X) \\
	\delta_{tdk} & = & P(Z_t=dk | S_t=d, X)
\end{eqnarray*}

\paragraph{E-Step.} As  $P(S, Z|X) = P(S|X) P(Z|S, X)$, the conditional distribution of the hidden variables can be calculated in two steps. First, $P(S|X)$ is the conditional distribution of the hidden Markovian state and can be calculated  via the forward-backward algorithm \citep[see][for further details]{Rabiner89} which only necessitates the current estimate of the transition matrix $\Pi^h$ and the current estimates of the emission densities at each observation point: $\psi^h_d(X_t)$. This algorithm provides the two conditional probabilities involved in the completed log-likelihood: $\tau_{td}$ and $\eta_{tdd'}$. Second, $P(Z_t|S_t=d, X)$ is given by
$$
\widehat{\delta}_{tdk}
	= \frac{\lambda_{dk} \phi(X_t, \gamma_{dk})}{\sum_{j=1}^{K_d} \lambda_{dj} \phi(X_t, \gamma_{dj})}.
$$

\paragraph{M-Step.} The maximization of the completed log-likelihood is straightforward and we get
\begin{eqnarray*}
	\widehat{\pi}_{dd'} & \propto & \sum_{t=1}^n \eta_{tdd'}, \\
    \widehat{\lambda}_{dk} & = & \frac{\sum_{t=1}^n \widehat{\tau}_{td} \widehat{\delta}_{tdk} }{\sum_{t=1}^n \widehat{\tau}_{td}}, \\
     \widehat{\gamma}_{dk} & = & \argmax_{\gamma_{dk}} \sum_{t=1}^n \widehat{\tau}_{td} \sum_{k=1}^{K_d} \widehat{\delta}_{tdk} \log \phi(x_t; \gamma_{dk}).
\end{eqnarray*}

\subsection{Hierarchical initialization of the EM algorithm}
\label{Methodology}

Like any EM algorithm, its behavior strongly 
depends on the initialization step. The naive idea of testing all the possible combinations of the
components leads to intractable calculations. We choose to follow the strategy proposed by \citep{Baudry2008}, which
is based on a hierarchical algorithm.
At each step of the hierarchical process, the best pair of clusters
$k,l$ to be combined is selected according to a criterion. To do this, 
we define three likelihood-based criteria adapted to the HMM context:
\begin{equation}
\label{ourICL}
\begin{array}{ll}
\nabla_{kl}^{X} &= \mathbb{E}_X\left[ \log P(X;G'_{k \cup l})\right] , \\
\nabla_{kl}^{X,S} &= \mathbb{E}_X\left[\log P(X,S;G'_{k \cup l})\right],\\
\nabla_{kl}^{X,Z}&= \mathbb{E}_X\left[\log P(X,Z;G'_{k \cup l})\right]
\end{array}
\end{equation}
where $G'_{k \cup l}$ is the $G-1$ clusters obtained by
merging the two clusters $k$ and $l$ from the model with $G$ clusters
($D<G<K$). It is assumed that the hierarchical algorithm is at the
$G$-th step and therefore the term `cluster' refers to either a
component or a mixture of components. Two clusters $k$ and $l$ are
merged if they maximise one of the merging criteria $\nabla_{kl}$:
\begin{eqnarray}
(k,l) = \argmax_{k,l \in \{1,...,G\}^2} \nabla_{kl}\;.
\label{ourICL2}
\end{eqnarray}

Once the two clusters $k$ and $l$ have been combined into a new
cluster $k'$, we obtain a model with $G-1$ clusters where the density
of the cluster $k'$ is defined by the mixture distributions of
clusters $k$ and $l$. Due to the constraints applied on the transition
matrix of ${Z_t}$, the resulting estimates of the model parameters do not correspond to the
ML estimates. To get closer to a local maximum, a few
iterations of the EM algorithm are proceeded to increase the likelihood of the
reduced model. The algorithm corresponding to the hierarchical
procedure described above is given in Appendix \ref{algo}.

\section{Selection Criteria for the number of hidden states} \label{SelectionCrit}

We recall that the choice of $K$ is left to the user. Given $K$ and $D$,
the EM algorithm is initialized by a hierarchical algorithm. In many 
situations, 
$D$ is unknown and difficult to choose. To tackle this problem, we propose
 model selection  criteria, derived from the classical mixture framework.

From a Bayesian point of view, the model $m \in \mathcal{M}$ 
maximizing the posterior probability $P(m|X)$ is to be chosen.
By Bayes theorem
\[
P(m|X)=\frac{P(X|m)P(m)}{P(X)},
\]
and supposing a non informative uniform prior distribution
$P(m)$ on the models of the collection, it leads to $P(m|X) \propto
P(X|m)$. Thus the chosen model satisfies
\[
\tilde{m}=\underset{m \in \mathcal{M}} \argmax P(X|m),
\]
where the integrated likelihood $P(X|m)$ is defined by
\[
P(X|m)=\int P(X|m, \theta) \pi(\theta|m)
d\theta,
\]
$\pi(\theta|m)$ being the prior distribution of the vector
parameter $\theta$ of the model $m$. Since this integrated 
likelihood is typically difficult to calculate, an asymptotic 
approximation of $2\ln\{P(X|m)\}$ is generally used. This approximation
is the Bayesian Information Criterion (BIC) defined by
\begin{eqnarray}
BIC(m) = \log P(X|m,\widehat{\theta}) - \frac{\nu_{m}}{2} \log (n).
\label{BIC2}
\end{eqnarray}
where $\nu_{m}$ is the number of free parameters of the model $m$ 
and $P(X|m,\widehat{\theta})$ is the maximum
likelihood under this model \citep{Schwarz1978}. 
Under certain conditions, BIC consistently estimates the number of mixture
groups \citep{Keribin2000}. But, as BIC is not
devoted to classification, it is expected to mostly select the
dimension according to the global fit of the model.
In the context of model-based clustering with a general latent variable
$U$, \citep{Biernacki2000} have
proposed to select the number of clusters based on 
the integrated complete likelihood $P(X,U|m,\theta)$
\begin{eqnarray}
P(X,U|m) = \int P(X,U |m,\theta) \pi(\theta|m) d\theta.
\label{ICLint}
\end{eqnarray}
A BIC-like approximation of this integral leads to the so-called
ICL criterion:
\begin{eqnarray}
ICL(m) = \log P(X,\widehat{U}|m,\widehat{\theta})  - \frac{\nu_m}{2} \log (n),
\label{ICL}
\end{eqnarray}
where $\widehat{U}$ stands for posterior mode of $U$. This definition
of ICL relies on a hard partitioning of the data and
\citep{McLachlan2000} proposed to replace $\widehat U$ with the
conditional expectation of $U$ given the observation and get
\begin{eqnarray}
ICL(m) &=& \E_X\left[\log
  P(X,U|m,\widehat{\theta})\right]-\frac{\nu_m}{2} \log (n) \label{ICLexp}
\\
&=&\log P(X|m,\widehat{\theta}) - \mathcal{H}_X(U) -
\frac{\nu_m}{2} \log (n). \nonumber
\end{eqnarray}
Hence, ICL is equivalent to BIC with an additional penalty
term, which is the conditional entropy of the hidden variable
$\mathcal{H}_X(U) = -\E_X\left[\log
  P(U|X, m, \widehat{\theta}) \right]$. This entropy is a measure of the
uncertainty of the classification. ICL is hence clearly dedicated to a
classification purpose, as it penalizes models for which the
classification is uncertain. One may also note that, in this context,
ICL simply amounts at adding the BIC penalty to the completed
log-likelihood, rather than to the log-likelihood
itself. ICL has been established in the independent mixture
context. Nonetheless, \citep{Celeux2008} used ICL in the HMM context 
and showed that it seems to have the same behaviour.

In our model, the latent variable $U$ is the couple $(S, Z)$, and 
a direct rewriting of \eqref{ICL} leads to
\begin{equation}
ICL(m) = \log P(X|m,\widehat{\theta}) - \mathcal{H}_X(S, Z)
- \frac{\nu_m}{2} \log (n)
\label{ICLZ}
\end{equation}
and the conditional entropy of $U = (S, Z)$ can further be
decomposed as
$$
\mathcal{H}_X(S, Z) = \mathcal{H}_X(S) + \mathbb{E}_X[\mathcal{H}_{X, S}(Z)],
$$ 
which gives raise to two different entropies: $\mathcal{H}_X(S)$
measures the uncertainty of the classification into the hidden states
whereas $\mathbb{E}_X[\mathcal{H}_{X, S}(Z)]$ measures the
classification uncertainty among the components, within each 
hidden states. This latter entropy may not be relevant for our purpose as
it only refers to a within-class uncertainty. We therefore propose to focus on the integrated complete likelihood $P(X,S|m,\theta)$ and derive 
an alternative version of ICL, where only the former entropy is used for
penalization, defined by:
\begin{equation}
ICL_S(m)  =  \log P(X|m,\widehat{\theta}) - \mathcal{H}_X(S)
- \frac{\nu_m}{2} \log (n).
\label{ICLS}
\end{equation}

These three criteria, $BIC, ICL$ and, $ICL_S$,
display different behavior in independent mixtures and in HMM. 
In the independent case, the number of free
parameters $\nu_m$ only depends on $K$ and the observed likelihood $\log
P(X|m, \widehat{\theta})$ remains the same for a fixed $K$, whatever
$D$. The $BIC$ and the $ICL$ given in Equations (\ref{BIC2}),
(\ref{ICLZ}) are thereby constant whatever the number of clusters.
Moreover, the $ICL_S$ always increases with
the number of clusters so none of these three criteria can be used in
the independent mixture context.  On the contrary, in the case of HMM,
the observed likelihood 
$\log P(X|m, \widehat{\theta})$ varies with the number of
hidden states. Furthermore, because the number free parameters 
 does depend on $D$ through the
dimension of the transition matrix, the number of free parameters of
a $D$-state HMM differs from that of a $(D-1)$-state HMM, even with
same $K$. This allows us the use of these three criteria
to select the number of clusters.

If we go back to the merging criteria of the hierarchical initialization of the EM algorithm defined in \ref{Methodology}. We note that each 
criterion $\nabla$ is related to one model selection criterion defined 
above. Indeed, the maximisation of $\nabla$ is equivalent
to:
\begin{eqnarray*}
\argmax_{k,l \in \{1,...,G\}^2} \nabla_{kl}^{X} &= & \argmin_{k,l}
\left[ BIC(G) - BIC(G'_{k \cup l}) \right]\;, \\
\argmax_{k,l \in \{1,...,G\}^2} \nabla_{kl}^{X,S} &= & \argmin_{k,l}
\left[ ICL_S(G) - ICL_S(G'_{k \cup l}) \right]\;,\\
\argmax_{k,l \in \{1,...,G\}^2} \nabla_{kl}^{X,Z} &= & \argmin_{k,l}
\left[ ICL(G) - ICL(G'_{k \cup l}) \right]\;.
\end{eqnarray*}

\section{Simulation studies}
\label{Simul}

In this section, we present two simulation studies to illustrate the performance of our approach. In Section \ref{Simul1}, we aim at determining the best combination between the selection criteria ($BIC$, $ICL_S$, $ICL$) and the merging criteria ($\nabla^{X}$, $\nabla^{X,S}$, $\nabla^{X,Z}$). In Section \ref{simul2}, we compare our method to that of \citep{Baudry2008} and we focus on the advantage of accounting for Markovian dependency in the combination process.

\subsection{Choice of merging and selection criteria}
\label{Simul1}
\subsubsection{Design}
\label{Design}
\begin{figure*}[ht!]
\centering \includegraphics[width=0.75\textwidth]{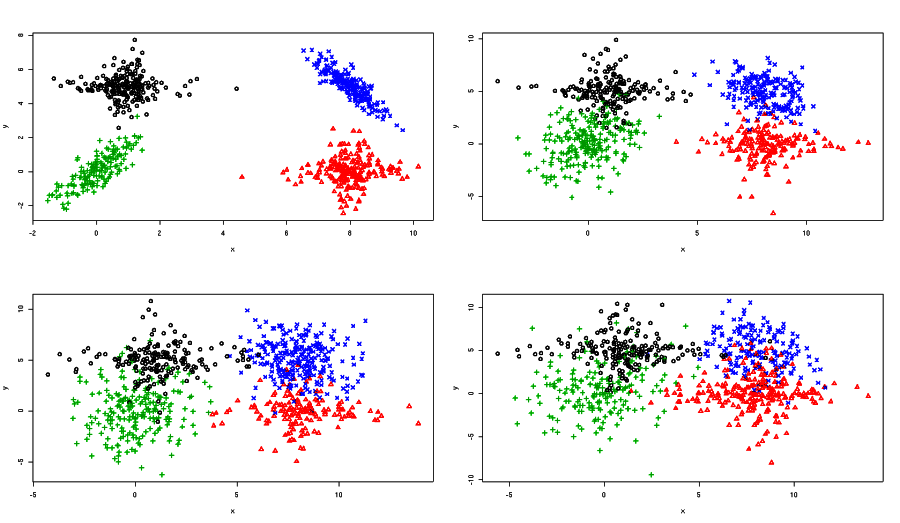}
 \caption{Example of simulated data. Top left:  $b=1$, easiest case, corresponds to the one studied by \citep{Baudry2008}; top right: $b=3$; bottom left: $b=5$; bottom right: $b=7$. Each group is represented by a symbol/color.}
\label{fig:1}
\end{figure*}

The simulation design is the same as that of \citep{Baudry2008}, with an additional Markovian dependency. We simulated a four-state HMM with Markov chain $\{S_t\}_t$. The emission distribution is a bidimensional Gaussian for the first two states and is a mixture of two bidimensional Gaussians for the other two. Therefore there are six components but only four clusters. In order to measure the impact of the Markovian dependency on posterior probability estimation, we considered four different transition matrices such that $\forall d \in \{1,...,4\}, \ P(S_t=d|S_{t-1}= d) = a$, with $a \in  \{0.25,0.5,0.75,0.9\}$ and $P(S_t=d'|S_{t-1}= d) = (1-a)/3$ for $d \neq d'$. The degree of dependency in the data decreases with $a$. To control the data shape, we introduce a parameter $b$ in the variance-covariance matrices $\Sigma_{k}$ of the $k$-th bidimensional Gaussian distributions such as $\Sigma_{k} = b \left(\begin{array}{cc} \sigma_{k 1} & \sigma_{k 12} \\ \sigma_{k 12} & \sigma_{k 2} \\ \end{array} \right)$. The parameter $b$ takes its values in $\{1,3,5,7\}$ where $b=1$ corresponds to well-separated clusters \citep[case of]{Baudry2008} and $b=7$ leads to overlapping clusters.  Figure \ref{fig:1} displays a simulated dataset for each value of $b$.
The mean and covariance parameter values are given in Appendix \ref{parameter}.

For each of the 16 configurations of $(a,b)$, we generated $C = 100$ simulated datasets of size $n = 800$. For the method we proposed, the inference of the $K$-state HMM has been made with spherical Gaussian distributions for the emission, i.e. the variance-covariance matrix is  $\left(\begin{array}{cc} \sigma^2 & 0 \\ 0 & \sigma^2 \\ \end{array} \right)$.

The performance of the method is evaluated by both the MSE (Mean Square Error) and the classification error rate.
The MSE of the conditional probabilities measures the accuracy of a method in terms of classification. It evaluates the difference between the conditional probability estimation $\widehat{\tau}^{(\nabla)}$ of a criterion $\nabla$ and the theoretical probability $\tau^{(th)}$:
\begin{eqnarray}
 MSE^{(\nabla)} = \frac{1}{C}\sum_{c=1}^{C} \frac{1}{n}\sum_{t=1}^n ||\widehat{\tau}_t^{(\nabla)}-\tau_t^{(th)}||_2.
\label{MSE}
\end{eqnarray}
The smaller the MSE, the better the performance.
Since our aim is to classify the data in a given number of clusters, another interesting indicator is the rate of correct classification. 
This rate allows the consistency of the classification to be measured, with respect to the true one. 
We calculated this rate for each simulation configuration where the classification has been obtained with the MAP (Maximum A Posteriori) rule.

\subsubsection{Merging criteria}
\label{Result}
\begin{figure*}[ht!]
\centering\includegraphics[width=0.75\textwidth]{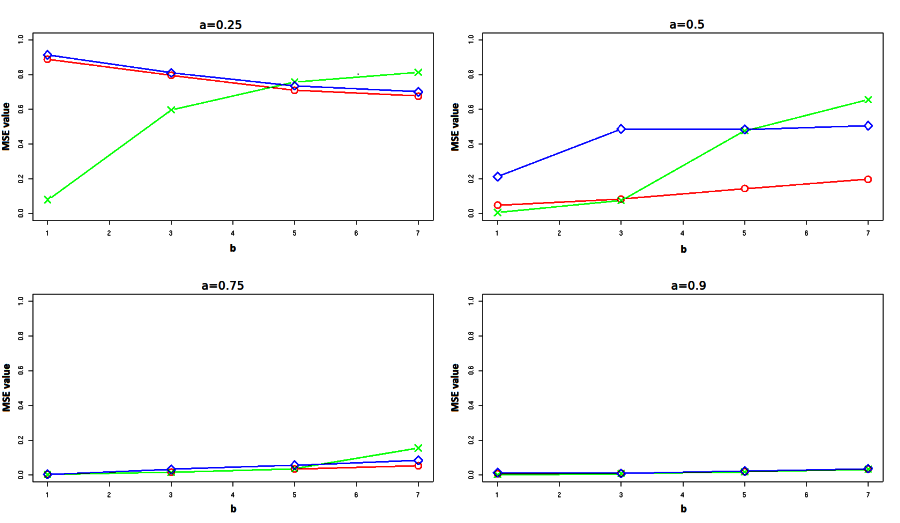}
 \caption{MSE values for each simulation condition. Each graph corresponds to a transition matrix which goes from low dependency level (top left) to high (bottom right). ``$\circ$'': $\nabla^{X}$, ``$\times$'': $\nabla^{X,S}$, ``$\Diamond$'': $\nabla^{X,Z}$.}
 \label{fig:2} 
\end{figure*}

The goal is to study the best way to merge the clusters of an HMM. Therefore, we compare the three merging criteria with regard to the MSE (see Figure \ref{fig:2}).

\paragraph{Dependency contribution.}
 When the data are independent ($a=0.25$), we note that the $\nabla^X$ and $\nabla^{X,Z}$ criteria provide a bad estimation whatever the value of $b$. The results obtained with $\nabla^{X,S}$ are  satisfying only if the groups are not overlapping ($b=1$). These poor results can be explained by the definition of the merging criteria that are not suited to the independent case.

From a general point of view, increasing the value of $a$ yields a better estimation of the conditional probabilities. When $a\neq0.25$, the $\nabla^{X}$ criterion provides the best results in most cases. In high dependency cases ($a = 0.75$ or $0.9$), the results are similar whatever the merging criterion. However, for $a = 0.5$, the $\nabla^{X,Z}$ generates estimations far from the true ones even if the groups are easily distinguishable ($b=1, 3$). Further simulation studies (not shown) point out that the $\nabla^{X}$ criterion outperformed the others as soon as $a \geq 0.4$.

\paragraph{Effect of the overlap.}

When $a = 0.5$, the criteria $\nabla^{X}$ and $\nabla^{X,S}$ produce similar estimates when the groups are well separated ($b = 1,3$). Increasing the overlap between the groups ($b= 5,7$) has very little influence on the results provided by $\nabla^{X}$ but is harmful to the $\nabla^{X,S}$. When the degree of dependency increases ($a = 0.75$ or $0.9$), the criterion $\nabla^{X}$ still gives the best results whatever the value of $b$.
  
\begin{figure*}[ht!]
\centering\includegraphics[width=0.75\textwidth]{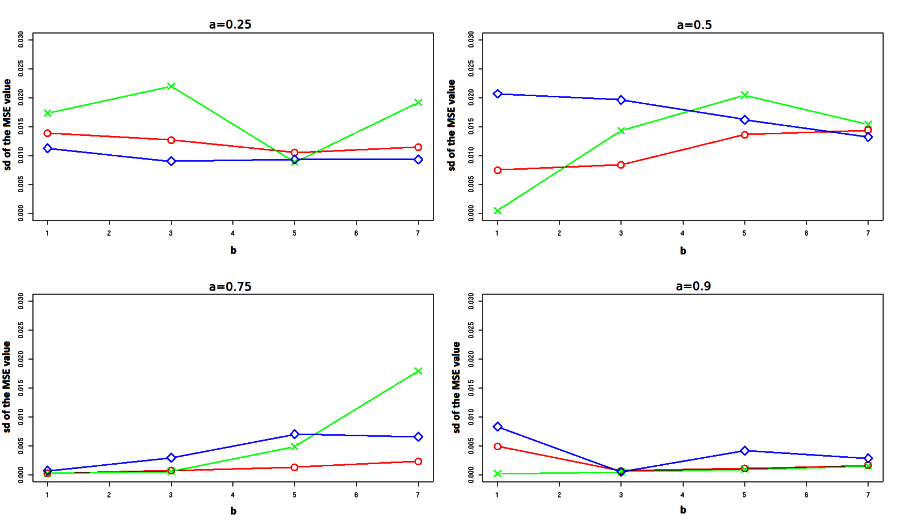}
 \caption{Standard deviation of the MSE for each simulation condition. Each graph corresponds to a transition matrix which goes from low dependency level (top left) to high (bottom right). ``$\circ$'': $\nabla^{X}$, ``$\times$'': $\nabla^{X,S}$, ``$\Diamond$'': $\nabla^{X,Z}$.}
\label{fig:3}
\end{figure*}

Figure \ref{fig:3} describes the variation in the standard deviation of the MSE for each simulation condition. Once more the $\nabla^{X}$ has the best results among the four methods, especially for high dependency level.\\
Table \ref{MisClass} shows the rate of correct classification with its standard deviation.
  \begin{table}
     \caption{Rate of correct classification and its standard deviation (mean(sd)). Top: low dependency. Bottom: high dependency. The values in bold correspond to the best rate of correct classification calculated for the three criteria.}
 \label{MisClass}
 \scriptsize
  \begin{center}
  \begin{tabular*}{0.55\textwidth}{lllll}
\hline\noalign{\smallskip}
   \multicolumn{2}{c}{\mbox{Parameters}} &  $\nabla^{X}$ & $\nabla^{X,S}$ & $\nabla^{X,Z}$ \\
  \hline
\multirow{4}{0.05cm}{\rotatebox{90}{a=0.25}}&b=1 & 0.479 (0.0076) & \textbf{0.958 (0.0087)} & 0.456 (0.0055)\\ 
  & b=3& 0.457 (0.0065) & \textbf{0.662 (0.0112)} & 0.427 (0.005)\\ 
  & b=5& 0.458 (0.0066) & \textbf{0.542 (0.0043)} & 0.422 (0.005)\\ 
  & b=7& 0.444 (0.0066) & \textbf{0.482 (0.0095)} & 0.409 (0.0055)\\ 
\noalign{\smallskip}\hline\noalign{\smallskip}
\multirow{4}{0.05cm}{\rotatebox{90}{a=0.5}} & b=1& 0.969 (0.0051) & \textbf{0.996 (0.0003)} & 0.871 (0.0122)\\ 
  & b=3& 0.913 (0.0054) & \textbf{0.928 (0.0071)} & 0.652 (0.0119)\\ 
  & b=5& \textbf{0.840 (0.0082)} & 0.694 (0.0106) & 0.615 (0.0104)\\ 
  & b=7& \textbf{0.776 (0.0089)} & 0.575 (0.0079) & 0.564 (0.0085)\\  
\noalign{\smallskip}\hline\noalign{\smallskip}
\multirow{4}{0.05cm}{\rotatebox{90}{a=0.75}} & b=1& \textbf{0.998 (0.0002)} & \textbf{0.998 (0.0002)} & 0.997 (0.0004)\\ 
  & b=3& 0.974 (0.0007) & \textbf{0.975 (0.0007)} & 0.962 (0.0021)\\ 
  & b=5& \textbf{0.940 (0.0013)} & \textbf{0.940 (0.0027)} & 0.926 (0.0041)\\ 
  & b=7& \textbf{0.907 (0.0020)} & 0.859 (0.0090) & 0.887 (0.0041)\\  
\noalign{\smallskip}\hline\noalign{\smallskip}
\multirow{4}{0.05cm}{\rotatebox{90}{a=0.9}} & b=1& 0.995 (0.0032) & \textbf{0.998 (0.0001)} & 0.991 (0.0052)\\ 
  & b=3& \textbf{0.989 (0.0005)} & \textbf{0.989 (0.0004)} & 0.989 (0.0005)\\ 
  & b=5& 0.974 (0.0010) & \textbf{0.975 (0.0009)} & 0.972 (0.0029)\\ 
  & b=7& \textbf{0.955 (0.0013)} & \textbf{0.955 (0.0012)} & 0.952 (0.0021)\\  
\noalign{\smallskip}\hline
  \end{tabular*}
\end{center}
   \end{table}
   \normalsize
When $a=0.25$, the $\nabla^{X,S}$ outperformed the other criteria whatever the value of $b$. For all other cases ($a\neq 0.25$), merging the components with $\nabla^{X,S}$ allows close or better results than those obtained by the $\nabla^{X}$ when $b$ either equals $1$ or $3$. However, when the case is more complicated ($b=5,7$), the $\nabla^{X}$ really outperforms. The $\nabla^{X,Z}$ provides the worst results among the three cases proposed. The difference between the methods is more flagrant on the MSE values (see Figure \ref{fig:2}). According to 
the above results, we propose the $\nabla^{X}$ criterion for merging 
the clusters.

\paragraph{Study of the combination of the merging and selection criteria.} We now focus on the estimation of the number of clusters, which equals $4$ for the simulation design given in Section \ref{Design}. We compare the three selection criteria proposed in Section \ref{SelectionCrit} and we study the estimated number of clusters for each simulation condition.
\begin{table}
    \caption{ Rate of correct estimations of the number of clusters for each simulation condition. The merging is done by the $\nabla^{X}$ criteria. Top: low dependency. Bottom:  dependency. The values in bold correspond to the best rate of correct classification calculated for the three methods.}
    \label{RateCluster}
  \begin{center}
  \scriptsize
 \begin{tabular*}{0.48\textwidth}{@{\extracolsep{\fill}}lllll}
\hline\noalign{\smallskip}
  \multicolumn{2}{c}{\mbox{Parameters}} &  BIC & $ICL_S$ & $ICL$\\
\noalign{\smallskip}\hline\noalign{\smallskip}
\multirow{4}{0.05cm}{\rotatebox{90}{a=0.25}} & b=1&  \textbf{0.88} &  0.80  & 0.69 \\ 
 & b=3&  \textbf{0.97}   & 0.85   & 0.38   \\ 
 & b=5& \textbf{0.98}   & 0.87   & 0.38         \\ 
 & b=7& \textbf{0.98}   & 0.92   & 0.21         \\ 
\noalign{\smallskip}\hline\noalign{\smallskip}
\multirow{4}{0.05cm}{\rotatebox{90}{a=0.5}} & b=1& 0.88   & \textbf{0.96}   & 0.49 \\ 
 & b=3& 0.89   & \textbf{0.98}   & 0.16  \\ 
 & b=5& 0.95   & \textbf{0.99}   & 0.12   \\ 
 & b=7& 0.96  & \textbf{1}    & 0.05  \\ 
\noalign{\smallskip}\hline\noalign{\smallskip}
\multirow{4}{0.05cm}{\rotatebox{90}{a=0.75}}&  b=1& 0.92 & \textbf{1}   & 0.62   \\ 
 & b=3& 0.96  & \textbf{1}   & 0.17  \\ 
 & b=5& 0.96  & \textbf{1}   & 0.15 \\ 
 & b=7&  0.97  & \textbf{1}   & 0.11\\ 
\noalign{\smallskip}\hline\noalign{\smallskip}
\multirow{4}{0.05cm}{\rotatebox{90}{a=0.9}}&  b=1&  0.92   & \textbf{0.96}   & 0.50 \\ 
 & b=3& 0.96  & \textbf{1}   & 0.29 \\ 
 & b=5& \textbf{1}  & \textbf{1}  & 0.26 \\ 
 & b=7& 0.98  & \textbf{1}   & 0.16   \\ 
\noalign{\smallskip}\hline
 \end{tabular*}
  \end{center}
 \end{table}
\normalsize
Table \ref{RateCluster} provides the rate of good estimations of the number of clusters. This rate is calculated for each dependency level and for each value of $b$.
 First of all, considering the $ICL$ as a selection criterion does not lead to a good estimation of the number of clusters. This can be explained by the fact that $ICL$ involves the latent variable $Z$ which is linked to the components. Hence, the $ICL$ tends to overestimate the number of clusters. It is more reliable to estimate the number of clusters with a criterion which does not depend on $Z$ such as the $BIC$ or the $ICL_S$. As shown in Table \ref{RateCluster} the best criterion for estimating the number of clusters is $ICL_S$.

\subsubsection{Conclusion}

We proposed three different criteria for combining the clusters and we showed that the $\nabla^{X}$ seems to outperform the other criteria when the aim is merging the components. In fact, it provides estimation of the conditional probabilities close to the true ones and these estimations are very robust in terms of MSE. Moreover, this is also confirmed by studying the rate of correct classification.
For the estimation of the number of clusters,  $ICL_S$ seems to be the most accurate. To conclude, we proposed using $\nabla^{X}$ as the merging criterion and estimating the number of clusters by $ICL_S$. Throughout the remainder of the paper, this strategy is called "HMMmix".

\subsection{Markovian dependency contribution}
\label{simul2}
In this second simulation study, by comparing the method proposed by \citep{Baudry2008} to HMMmix, we are interested in taking into account the Markovian dependency.\\
For computing the \citep{Baudry2008} approach, we used the package Mclust \citep{Fraley1999} to run the EM algorithm for the estimation of the mixture parameters.\\
In the independent case ($a=0.25$), the method of \citep{Baudry2008} provides better estimation of the conditional probabilities than does the HMMmix (see Table \ref{MSE_BIC_Baudry}). The proposed method tries to find non-existent Markovian dependency, making it less efficient.

\begin{table}
  \caption{MSE values (mean(sd)) obtained with our method HMMmix and the method of \citep{Baudry2008}} 
  \label{MSE_BIC_Baudry}
  \scriptsize
  \begin{center}
  \begin{tabular*}{0.48\textwidth}{@{\extracolsep{\fill}}llll}
\hline\noalign{\smallskip}
  \multicolumn{2}{c}{Parameters} &  HMMmix & Baudry \textit{et al.} (2008) \\
\noalign{\smallskip}\hline\noalign{\smallskip}
\multirow{4}{0.05cm}{\rotatebox{90}{a=0.25}}
  &b=1 & 0.887 (0.014) & \textbf{0.005 (0.0087)}\\ 
  & b=3& 0.795 (0.013) & \textbf{0.144 (0.022)}\\ 
  & b=5& 0.709 (0.011) & \textbf{0.319 (0.022)} \\ 
  & b=7& 0.676 (0.011) & \textbf{0.358 (0.017)}\\ 
  \noalign{\smallskip}\hline\noalign{\smallskip}
\multirow{4}{0.05cm}{\rotatebox{90}{a=0.5}}
   & b=1& 0.048 (0.007) & \textbf{0.006 (0.005)}\\ 
  & b=3& \textbf{0.084 (0.008)} & 0.169 (0.022)\\ 
  & b=5& \textbf{0.144 (0.014)} & 0.385 (0.021) \\ 
  & b=7& \textbf{0.198 (0.014)} & 0.369 (0.018) \\  
  \noalign{\smallskip}\hline\noalign{\smallskip}
\multirow{4}{0.05cm}{\rotatebox{90}{a=0.75}} & b=1& \textbf{0.003 (0.0002)} & 0.013 (0.007)\\ 
  & b=3& \textbf{0.016 (0.0007)} & 0.173 (0.022) \\ 
  & b=5& \textbf{0.035 (0.001)} & 0.407 (0.020) \\ 
  & b=7& \textbf{0.054 (0.002)} & 0.421 (0.017) \\   
\noalign{\smallskip}\hline\noalign{\smallskip}
\multirow{4}{0.05cm}{\rotatebox{90}{a=0.9}}  & b=1& \textbf{0.007 (0.005)} & \textbf{0.003 (0.0002)}\\ 
  & b=3& \textbf{0.009 (0.0007)} & 0.229 (0.023)\\ 
  & b=5& \textbf{0.019 (0.001)} & 0.413 (0.022) \\ 
  & b=7& \textbf{0.034 (0.002)} & 0.418 (0.017) \\  
\noalign{\smallskip}\hline
  \end{tabular*}
 \end{center}
 \end{table}
\normalsize

Note that, whatever the value of $a$, the method of \citep{Baudry2008} logically provides the same results. Regarding the proposed method, we find that the Markovian dependency (when it exists) is beneficial for the estimation of conditional probabilities. The interest of accounting for dependency in the hierarchical process stands out for the more complicated configuration, i.e. when the groups are overlapping ($b \neq 1$). In this case, our method tends to be more robust.

\subsubsection{Two nested non-Gaussian clusters}
\begin{figure*}[ht!]
  \centering \includegraphics[width=0.5\textwidth,height=6cm]{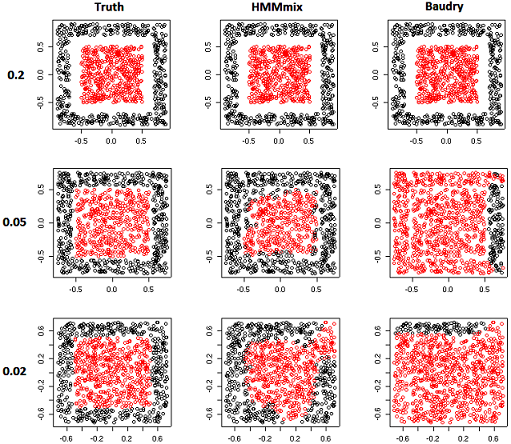} 
   \caption{The distance between the two squares goes from $0.2$ (top) to $0.02$ (bottom). Left: Truth, Middle: HMMmix, Right: Method of \citep{Baudry2008} }
   \label{fig:4}
  \end{figure*}

In this section, we focus on an HMM where the emission distribution is neither a Gaussian nor a mixture of Gaussian. We simulated datasets according to a binary Markov chain $S$ with transition matrix $\left(\begin{array}{cc} 2/3 & 1/3 \\ 1/3 & 2/3 \\ \end{array} \right)$. Denote by $C_a = [-a;a]^2$ the square of side length $2a$. The two clusters are nested (see Figure \ref{fig:4}) and correspond to the random variable $X$ such as:
\begin{itemize}
 \item[$\bullet$] $(X_t|S_t=0) \sim \mathcal{U}_{\mathcal{D}_0}$, with  $\mathcal{D}_0 = C_{\frac{1}{2}}$.
 \item[$\bullet$] $(X_t|S_t=1) \sim \mathcal{U}_{\mathcal{D}_1}$, with $\mathcal{D}_1 = C_{b+0.2} \backslash C_b$ and $b\in\{0.7,0.55,0.52\}$.\\
\end{itemize}
The parameter $b$ represents the distance between the two groups: $b\in\{0.02,0.05,0.2\}$. According to this simulation design, we simulated three different datasets (see Figure \ref{fig:4}).
Figure \ref{fig:4} displays the classification given by our method and the one given by \citep{Baudry2008}, according to the distance between the two squares.
Note that the spatial dependence cannot be observed on this figure. With our approach, the two nested clusters are well detected with a low number of misclassified which logically increases when the distance decreases.
If Markovian dependency is not taken into account \citep[method of][]{Baudry2008}, the two clusters are identified only when they are well separated. In such a complex design, geometrical considerations are not sufficient to detect the two clusters; accounting for dependency is therefore required. 

\section{Illustration on a real dataset}
\label{realdata}

We now consider the classification of probes in tiling array data from the model plant \textit{Arabidopsis thaliana}.
This experiment has been carried out on a tiling array of about 150 000 probes per chromosome. The biological objective of the experiment is to compare the methylation of a specific protein (histone H3K9) between a wildtype and the mutant 
$ddm1$ of the plant. It is known that the over-methylation or under-methylation of this protein is involved in the regulation process of gene expression.
As two adjacent probes cover the same genomic region it is required to take into account the dependency in the data. Due to computational time, which will be discussed in Section \ref{Discussion}, we apply our method on a sub-sample of 5000 probes of Chromosome 4.
\begin{figure*}[ht!]
  \centering \includegraphics[width=0.75\textwidth]{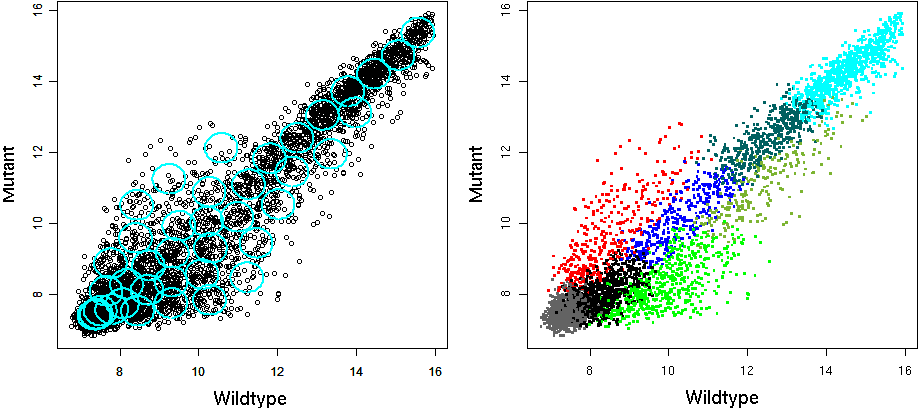}
 \caption{Left: Representation of the initial HMM with $K=40$ components. Right: Classification obtained after merging the components. Each color represents a specific cluster.}
\label{fig:5}
\end{figure*}
We apply the HMMmix method starting with $K=40$ components (see Figure \ref{fig:5}, Left). The number of clusters given by $ICL_S$ is $8$. Figure \ref{fig:5} (Right) displays the final classification. The cluster on the bottom left (in grey) represents the noise, \textit{i.e.} the unmethylated probes. The four groups on the diagonal correspond to the probes which have the same level of methylation for the two conditions. These probes have either  high (cyan) or low level (black) of methylation. The cluster on the left side of the diagonal (red) contains the over-methylated probes in the mutant compared to the wildtype, whereas the two clusters on the right side (green) correspond to the under-methylated ones.\\
The estimated densities are represented in Figure \ref{fig:6} for each cluster. The histograms are built by projecting the data on the X-axis (corresponding to the wildtype) weighted with their posterior probabilities. We see that the empirical distributions are not unimodal. Considering a mixture of distributions clearly leads to a better fit than single Gaussian. 
\begin{figure*}[ht!]
\centering\includegraphics[width=0.75\textwidth]{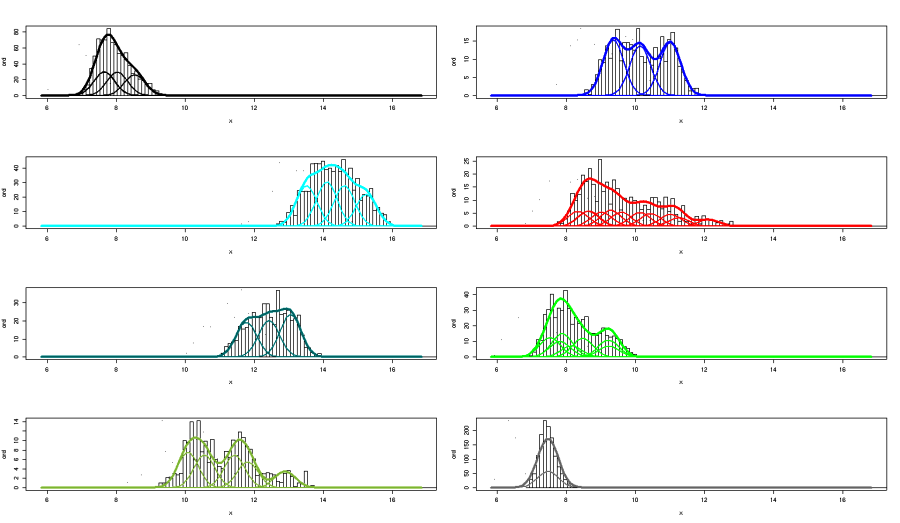} 
 \caption{Representation of the fit of densities for each cluster. The colors of the estimated densities correspond to the colors of the clusters presented in Figure \ref{fig:5}. }
\label{fig:6}
\end{figure*}

The proportions of the over-methylated and under-methylated clusters are $6.5\%$ and $15.5\%$, respectively. This result seems to be consistent with the biological expectation \citep{Lippman2004}. Moreover, in this dataset, there is one well-known transposable element named META1 which is an under-methylated region. With our method, $75\%$ of the probes of META1 are declared under-methylated.

In conclusion, the final classification is easily interpretable with respect to the biological knowledge. Furthermore, the flexibility of our model makes it possible to better fit the non-Gaussian distribution of real data.

\section{Discussion}
\label{Discussion}
In this article, we have proposed an HMM with mixture of parametric distributions as emission. This flexible modeling provides a better estimation of the cluster densities. Our method is based on a hierarchical combination of the components which leads to a local optimum of the likelihood. We have defined three criteria and in a simulation study we have highlighted that the $\nabla^{X}$ is the best criterion for merging components and $ICL_S$ for selecting the number of clusters. A real data analysis allows us to illustrate the performance of our method. We have shown that the clusters provided by our approach are consistent with the biological knowledge.
Although the method is described with an unknown hidden states $D$ and is illustrated with mixture of Gaussian distributions, we point out that tha same 
approach can be used when $D$ is known or with other parametric distribution 
family in the mixtures. For the initial number of components $K$, a 
brief simulation study has shown that for a large enough value of $K$, the classification still remains the same. 

A remaining problem of HMMmix is the computational time, especially when the size of the dataset is greater than $10 000$. This is due to the calculation of the $\nabla^{X}$ criterion which is linked to the observed log-likelihood. The computation of this observed log-likelihood requires the forward loop of the forward-backward algorithm whose complexity is linear in the number of observations. Otherwise, the number of models considered in the hierarchical procedure is 
$\sum_{d=D}^K d(d-1)/2 $and the computational time dramatically increases with $K$ and is of order $O(nK^3)$ . 
Consequently, to decrease the computational time, the solution is to 
reduce the space of models to explore. 
This can be done by a pruning criterion based on an approximation of $\nabla^{X}$ leading to a complexity at $O(nK^2)$

\bibliographystyle{plainnat}
\bibliography{RefComb} 

\begin{appendix}
\section*{Appendix}
\section{Algorithm}
\label{algo}
We present in this appendix the algorithm we proposed for merging components of an HMM. This algorithm has been written with respect to the results we obtained in Section \ref{Result}. However, this algorithm can easily be written for other criteria.

\begin{enumerate}
 \item Fit an HMM with $K$ components. 
\item From $G= K, K-1, ..., 1$
\begin{itemize}
 \item Select the clusters $k$ and $l$ to be combined as:

\begin{eqnarray*}
(k,l) = \argmax_{i,j \in \{1,...,K\}^2} \nabla_{ij}^{X},
\end{eqnarray*}

\item Update the parameters with a few steps of the EM algorithm to get closer to a local optimum.

\end{itemize}

\item Selection of the number of groups $\widehat{D}$:
\begin{eqnarray*}
\hat{D} &= &\argmax_{k\in\{K,...,1\}}ICL_S(k)\\ 
&=& \argmax_{k\in\{K,...,1\}} \log P(X,\widehat{S}|k,\widehat{\theta}_k)  - \frac{\nu_k}{2} \log (n),
\end{eqnarray*}
\end{enumerate}

 \section{Mean and variance of the Gaussian distributions for the simulation study (Section \ref{Simul1})}
 \label{parameter}
$$\mu_{1} = \left(\begin{array}{c} 1\\5 \end{array} \right), \mu_{2} = \left(\begin{array}{c} 1\\5 \end{array} \right), \mu_{3} = \left(\begin{array}{c} 8\\0 \end{array} \right),$$\\
$$ \mu_{4} = \left(\begin{array}{c} 8\\0 \end{array} \right), \mu_{5} = \left(\begin{array}{c} 0\\0 \end{array} \right), \mu_{6} = \left(\begin{array}{c} 8\\5 \end{array} \right).$$
and,

 $$\Sigma_{1} = \left(\begin{array}{cc} 0.1& 0\\ 0 & 1 \\ \end{array} \right), \Sigma_{2} = \left(\begin{array}{cc} 1 & 0 \\ 0 & 0.1\\ \end{array} \right), \Sigma_{3} = \left(\begin{array}{cc} 0.1 & 0 \\ 0 & 1\\ \end{array} \right),$$\\
 $$ \Sigma_{4} = \left(\begin{array}{cc} 1 & 0 \\ 0 & 0.1\\ \end{array} \right), \Sigma_{5} = \left(\begin{array}{cc} 0.4 & 0.5 \\ 0.5 & 1\\ \end{array} \right), \Sigma_{6} = \left(\begin{array}{cc} 0.3 & -0.4 \\ -0.4 & 0.7\\ \end{array} \right).$$

\end{appendix}
\end{document}